\def\year{2022}\relax
\documentclass[letterpaper]{article} 
\usepackage{aaai22}  
\usepackage{times}  
\usepackage{helvet}  
\usepackage{courier}  
\usepackage[hyphens]{url}  
\usepackage{graphicx} 
\usepackage{natbib}  
\usepackage{caption} 
\DeclareCaptionStyle{ruled}{labelfont=normalfont,labelsep=colon,strut=off} 
\usepackage{amsmath}
\usepackage{graphicx}
\usepackage{booktabs}
\usepackage{amssymb}
\usepackage{algorithm}
\usepackage{algorithmic}

\usepackage{newfloat}
\usepackage{listings}
\floatstyle{ruled}
\newfloat{listing}{tb}{lst}{}
\floatname{listing}{Listing}
\title{Learning to Select the Next Reasonable Mention for Entity Linking}
\author{
    Jian Sun, \textsuperscript{\rm 1,2}
   Yu Zhou, \textsuperscript{\rm 1,2,3}
   Chengqing Zong, \textsuperscript{\rm 1,2}
}
\affiliations{
    \textsuperscript{\rm 1} School of Artificial Intelligence, University of Chinese Academy of Sciences,\\
    Beijing 100049, P.R.China \\
    \textsuperscript{\rm 2} National Laboratory of Pattern Recognition, Institute of Automation, CAS \\
    \textsuperscript{\rm 3}Beijing Fanyu Technology Co., Ltd, Beijing, China\\
jian.sun@nlpr.ia.ac.cn, yzhou@nlpr.ia.ac.cn, cqzong@nlpr.ia.ac.cn

}

\usepackage{bibentry}

\begin{document}

\maketitle

\begin{abstract}
Entity linking aims to establish a link between entity mentions in a document and the corresponding entities in knowledge graphs (KGs). Previous work has shown the effectiveness of global coherence for entity linking. However, most of the existing global linking methods based on sequential decisions focus on how to utilize previously linked entities to enhance the later decisions. In those methods, the order of mention is fixed, making the model unable to adjust the subsequent linking targets according to the previously linked results, which will cause the previous information to be unreasonably utilized. To address the problem, we propose a novel model, called DyMen, to dynamically adjust the subsequent linking target based on the previously linked entities via reinforcement learning, enabling the model to select a link target that can fully use previously linked information. We sample mention by sliding window to reduce the action sampling space of reinforcement learning and maintain the semantic coherence of mention. Experiments conducted on several benchmark datasets have shown the effectiveness of the proposed model.
\end{abstract}

\section{Introduction}

As a critical step in the task of natural language understanding, entity linking (EL) has been widely used in many fields, such as question answering \cite{bordes2015large}, knowledge fusion \cite{jhyan2020driven}, and information extraction \cite{hoffmann2011knowledge}. This task is challenging because mentions are usually ambiguous, and it is difficult to link with the entities in the knowledge base only from the surface of mention. 
\begin{figure}[t] 
\centering
\includegraphics[width=7cm]{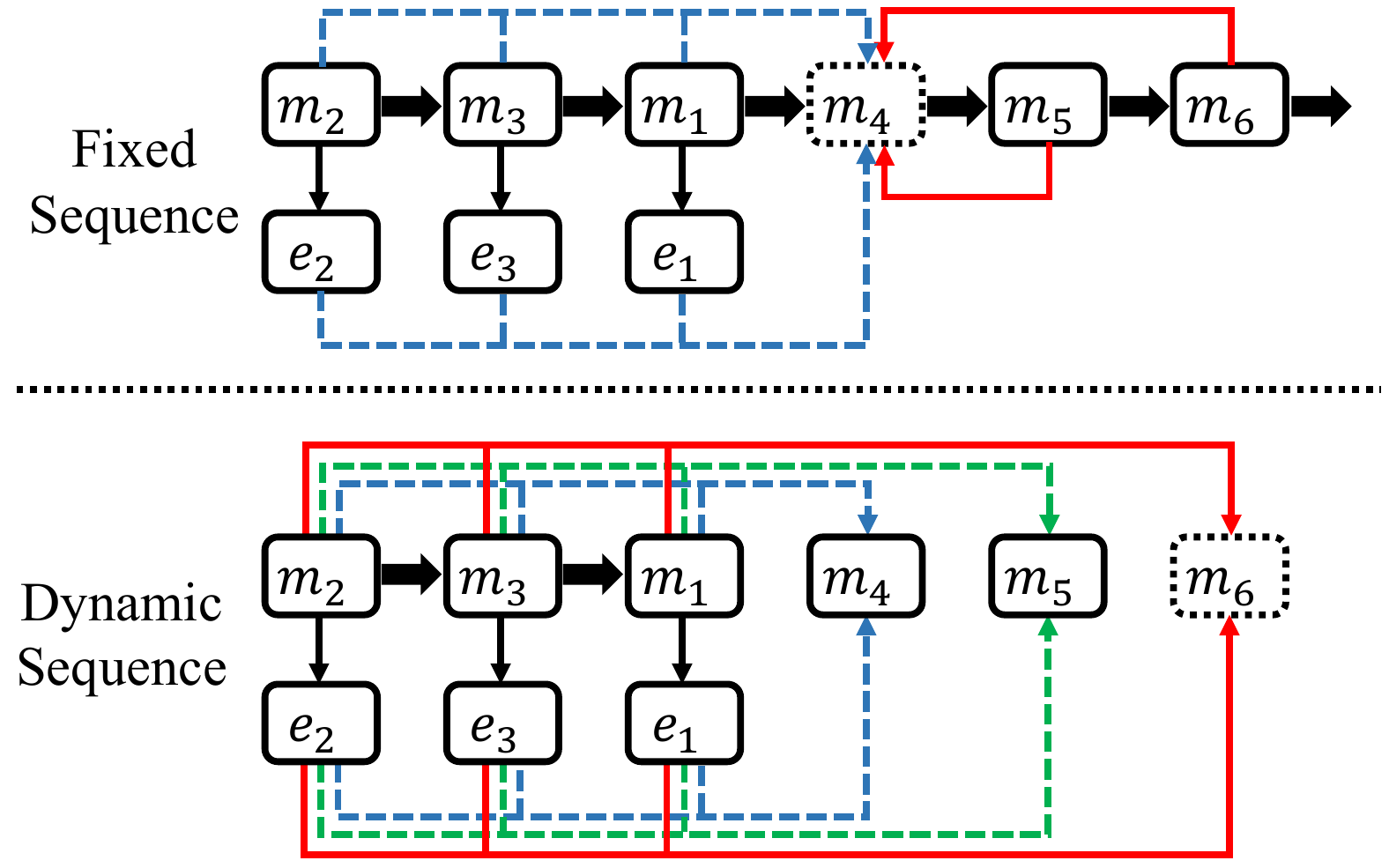}
\caption{
A comparative example of fixed sequence and dynamic sequence.}
\label{figure1:example} 
\end{figure}

To alleviate this problem, various EL models have been proposed \citep{huang_improving_2020,yang_learning_2019,zhang_joint_2020,ganea_deep_2017,le_improving_2018,cao_neural_2018}, which can be divided into local model and global model. The local model mainly focuses on contextual words around the mention, and the global model concerns the topic coherence.
Some global models \cite{ganea_deep_2017,le_improving_2018} calculate the pairwise coherence between all candidate entities, which suffers from high computational complexity and introduces some noises \cite{yang_learning_2019}. Therefore, some studies convert the global linking into a sequential decisions problem \cite{yang_learning_2019,fang_joint_2019,gu_read_2021}, which use previous information to disambiguate subsequent mentions. These studies use the natural order in the original document, the similarity between mentions, or the difficulty of entity linking to sort mentions and get a fixed sequence, on which a reinforcement learning algorithm or supervised learning method is used to link the entity.
However, the above fixed sequence is problematic. For example, in Figure \ref{figure1:example}, $m_i$ is a mention in the document, and $e_i$ is a candidate entity that has already been linked to $m_i$. $m_4$, $m_5$, and $m_6$  are the mentions to be linked. In the fixed sequence, $m_4$  is the next target to be linked, but sometimes $m_1$, $m_ 2$, and $m_ 3$ do not provide enough critical information to disambiguate $m_4$. On the contrary, the critical information used to disambiguate $m_4$ may be $m_5$ or $m_6$. Meanwhile, although $m_1$, $m_ 2$, and $m_ 3$ cannot be used to disambiguate $m_4$, they may be enough to disambiguate $m_6$. Obviously, the methods based on fixed sequence can not deal with the above situation.


 In this paper, to solve the above problems, we propose a global entity linking model where mention can be selected dynamically based on the results of previously linked entities. Instead of deciding which key mention can be used to disambiguate the current mention, we use the previously linked mention and entity information to select a target that can make the most of the previous information.
For example, in the dynamic sequence in Figure \ref{figure1:example}, $m_2$, $m_3$, and $m_1$ are the mentions that have been linked. 
Instead of choosing which mentions can be used to disambiguate $m_4$, we choose those that can make full use of $m_2$, $m_3$, and $m_1$ for disambiguation, such as $m_6$.
We utilize a policy network to choose a linking target that can fully use previously linked information. Through a sliding window, the model can sample a fixed number of mentions, which is beneficial to reducing the action search space of reinforcement learning and maintaining a certain semantic coherence between the mentions. To better explore the importance of the order of mention, we optimize our model from three different perspectives. We conduct experiments on different datasets, and the results show that our model outperforms state-of-the-art systems.

\section{Related Work}
Entity linking has gained increasing attention in recent years, which can be divided into two major models: local and global models. The local model utilizes the information around mention to disambiguate mentions. However, sometimes the context information is sparse, and it is difficult to make use of this sparse information to complete entity disambiguation \cite{yang_learning_2019}. Because of this, many global disambiguation models are proposed from the perspective of topic coherence
Some studies \cite{ganea_deep_2017,le_improving_2018,chen2020improving} assume that the target entities of all mentions in a document shall be related and utilize fully-connected pairwise Conditional Random Field model and exploit loopy belief propagation to estimate the max-marginal probability. This assumption is unreasonable. Sometimes, some entities in a document are unrelated to each other. 
With the popularity of graph neural networks, some studies \cite{wu2020dynamic,cao_neural_2018} apply graph neural networks (GNN) to encode mention-entity graphs. These approaches need to build a subgraph for all candidate entities of mentions and apply graph neural networks to integrate both local contextual features and global coherence information for entity linking. A high-quality subgraph is a key to the linking method based on GNN. Some studies \cite{yang_learning_2019,fang_joint_2019,gu_read_2021} convert the global linking into a sequential decisions problem. The core idea is to use previously linked entities to enhance the future mention disambiguation. 
The limitation of these methods is that the order of mention sequence is fixed while selecting the entity, and the reasonable link objects can not be selected dynamically according to the previous information, which will cause the previous information can not be used reasonably. Moreover, it can easily lead to error accumulation. In an extreme case, most of the previously linked results are incorrect, and these incorrect entities may negatively affect future link steps. Intuitively, the order of mention is related to the performance of subsequent entity linking and is the basis of entity linking.
\begin{figure}[t] 
\centering
\includegraphics[width=7cm]{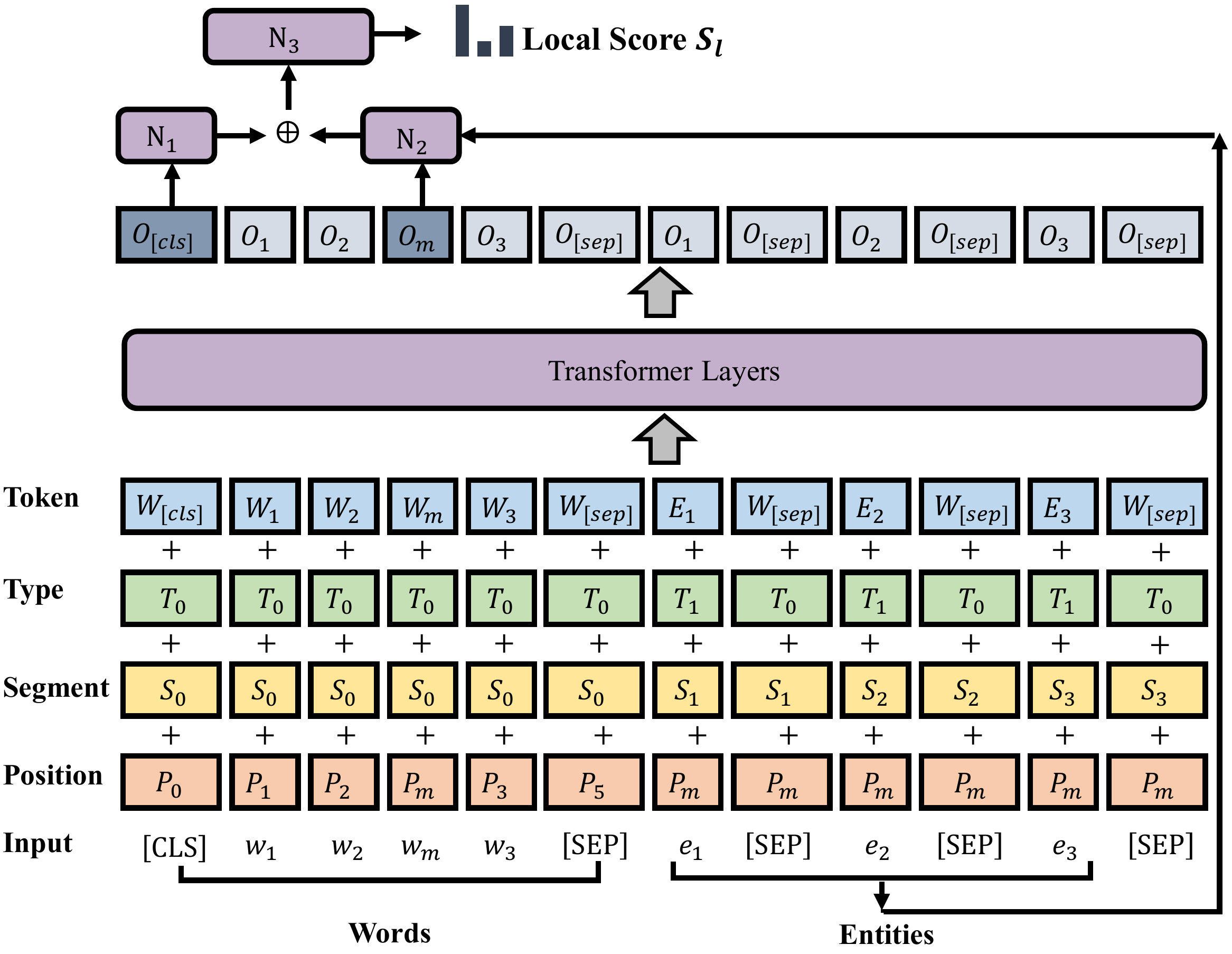}
\caption{The architecture of local entity linking model. $w_m$ is the mention and $w_i$ is the word in context surrounding mention $w_m$.}
\label{figure1:1} 
\end{figure}

\section{Methodology}
\subsection{Task Definition}
Formally, given a set of entity mentions $M = \{m_1, m_2,..., m_t\}$ in a document $D$, each mention $m_t \in D$ has a set of candidate entities $C_t = \{ e^1_t, e^2_t,..., e^n_t \}$. Entity linking maps each mention $m_t$ to the correct entity. Our entity linking model consists of two models: local model and global model. The local model only utilizes the context of mentions to disambiguate mentions. In contrast, the global model uses previously linked mentions to disambiguate subsequent mentions according to the topic consistency between mentions.
\subsection{Local Model}
In this section, we apply two different local models to the entity linking model. The first is the local model in \cite{ganea_deep_2017}, which measures the relevance of entity candidates of each mention independently. The local score function is as follows:
\begin{equation}
\Psi\left(e^i_t, \hat{w}_{t}\right)={\mathbf{e}^i_t}^{\top} \mathbf{B}_1 f\left(\hat{w}_{t}\right)
\end{equation}
where $\mathbf{e}^{i}_t$ is the embedding of entity $e^i_t$, $\mathbf{B}_1$ is a diagonal matrix and $f\left(\hat{w}_{t}\right)$ is the feature representation of local context surrounding $m_t$, $\hat{w}_t$ is the set of local context words.

We also design a new local entity linking model based on Transformer \cite{vaswani2017attention} encoder, as shown in Figure \ref{figure1:1}, which utilizes the mechanism of attention. Inspired by the popular pre-train language model such as BERT \cite{DBLP:journals/corr/abs-1810-04805}, we introduce two special symbols $\rm[CLS]$ and $\rm[SEP]$ to construct the input data. The special symbol $\rm[CLS]$ is utilized to encode the information of the entire sequence by self-attention, and the symbol $\rm[SEP]$ is used to distinguish between words and entities. 
Similar to the methods in BERT, we concatenate local context $w_t$ and candidate entities $\{e^1_t,e^2_t,...,e^n_t\}$ with $\rm[CLS]$ and $\rm[SEP]$ tokens as the input sequence in the following format:
\begin{equation}
{\rm[CLS]}\ w_1, w_2,...,w_t\ {\rm[SEP]}\ {e}_1 \ {\rm[SEP]},...,{e}_n\ {\rm[SEP]}
\end{equation}
where $w_t \in \hat{w_t}$ is the word in the local context, and $e_i \in \{{e}^1_t,{e}^2_t,...,{e}^n_t\}$ is the candidate entity. The input representation is constructed by summing up the following four parts:

(1) \textbf{Token embedding} represents the embedding of the word or entity. We represent the embedding of word and entity as $W$ and $E$, respectively. Here, $E_i=(w_{e_i}+\mathbf{B}_2\mathbf{e}_i)/2$, $w_{e_i}$ is the surface form of the entity and the $\mathbf{B}_2$ is to migrate entities from entity space to word space.
(2) \textbf{Type embedding} $T_k$ is used to distinguish the type of token. There are two types, words type $T_0$ and entity type $T_1$.
(3) \textbf{Segment embedding} $S_l$ is used to distinguish different candidate entities and contexts, where $l \in \{0,1,...,n\}$.
(4) \textbf{Position embedding} $P$ represents the position of the token in context. It is worth noting that there is no clear sequence relationship between different candidate entities. So, we use the position embedding of mention in context to represent the position embedding of all candidate entities. In this way, candidate entities can capture the same position information as mention.
We feed the above four parts into Transformer layers and get the output of words. We use ${\mathbf N_3}$ to get the result.
\begin{equation}
\begin{aligned}
{\mathbf N_1}&= \sigma(f_2(Drop(ReLU(f_1({\mathbf{o}_{cls}})))))\\
{\mathbf N_2}&= [\mathbf{e}_1,\mathbf{e}_2,...,\mathbf{e}_n] {\mathbf{o}}^\top_m \\
{\mathbf N_3}&= \sigma(f_4(Drop(ReLU(f_3([{\mathbf N_1};{\mathbf N_2}]))))) 
\end{aligned}
\end{equation}
where $\mathbf{o}_{cls}$ is the output representation of $\rm[CLS]$. $f_i$ is the fully connected layer. $\sigma$ is $\rm softmax$ function. $Drop(\cdot)$ is the dropout operation \cite{srivastava_dropout_2014}. $ReLU(\cdot)$ is the rectifier nonlinearity \cite{nair2010rectified}. $\mathbf N_1$, $\mathbf N_2$, and ${\mathbf N_3 \in \mathbb{R}^{1 \times n}}$. Finally, we get the local score $\Psi_{m}={\mathbf N_3}$.
\subsection{Global Model}
\begin{figure}[t] 
\centering
\includegraphics[width=8cm]{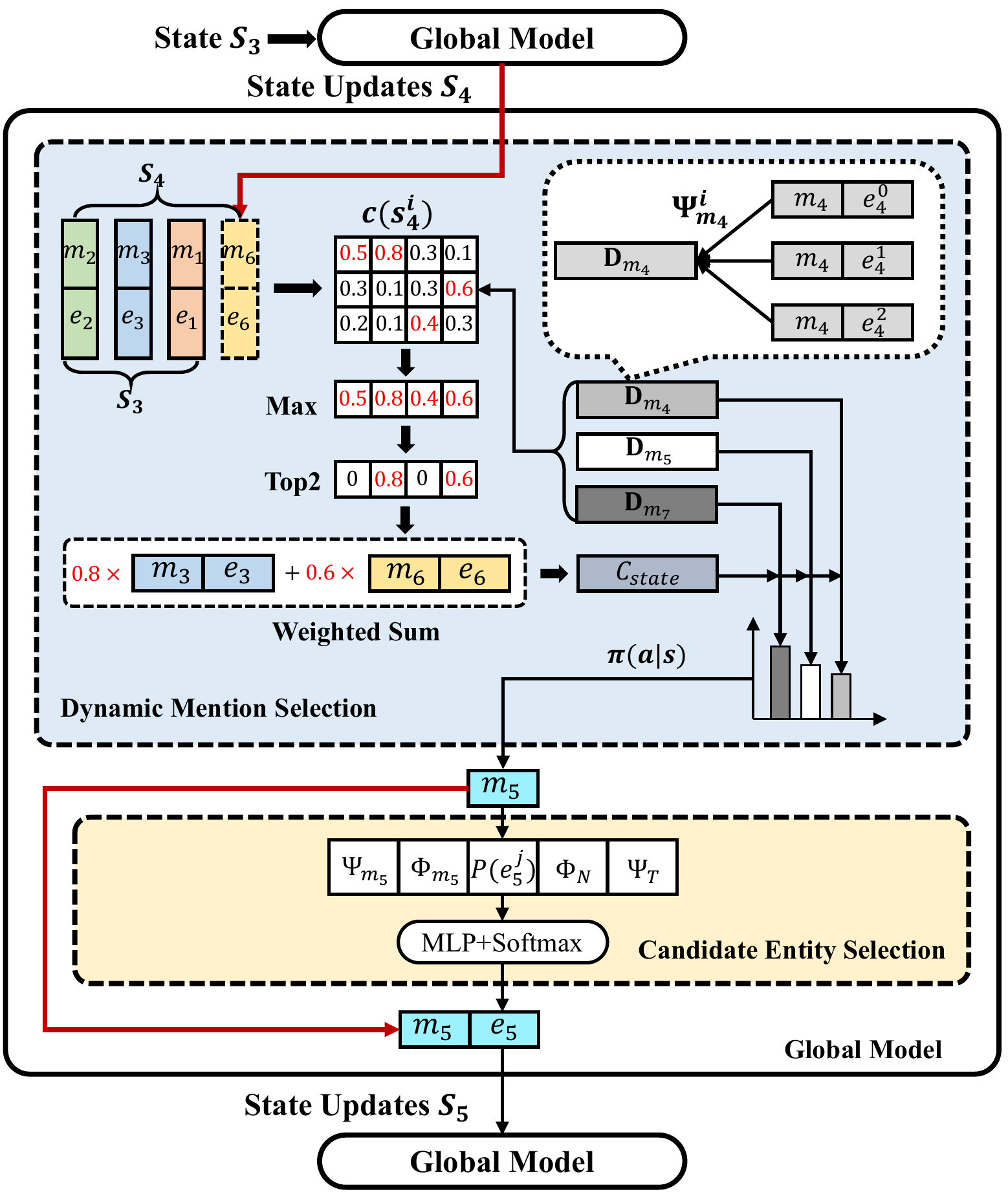}
\caption{The architecture of global model.}
\label{architecture} 
\end{figure}

Our global entity linking model consists of the dynamic mention selection model and the candidate entity selection model. The dynamic mention selection model utilizes the policy networks to select the corresponding mention, and the candidate entity selection model maps the mention to the correct target entity. The architecture of the global model is shown in Figure \ref{architecture}. 

\subsubsection{Dynamic Mention Selection}

The basic idea of dynamic mention selection is to use information (state) about the previously selected mention and the linked entities to select a reasonable mention from unresolved mentions. We consider the dynamic mention selection model as an agent. The agent utilizes the policy network to learn a stochastic policy and select an action (mention) according to the policy at each state. The stochastic policy can ensure the agent can explore the action space. Now, we will introduce the detail of how to select the target mention.

\noindent {\bfseries{States.}} We treat the previously selected mentions and entities as state information. Specifically, each state $S_t = \{ [m_0;e_0],[m_1;e_1],...,[m_t;e_t]\}$ is a set of mentions and entities, where $[m_1;e_1]$ represent the concatenation of the resolved mention $m_1$ and linked entity $e_1$. Note $m_1$ indicates the first mention selected according to the dynamic selection model, not the first mention appearing in the text. $m_0$ and $e_0$ are initial states.

\begin{figure}[ht] 
\centering
\includegraphics[width=6cm]{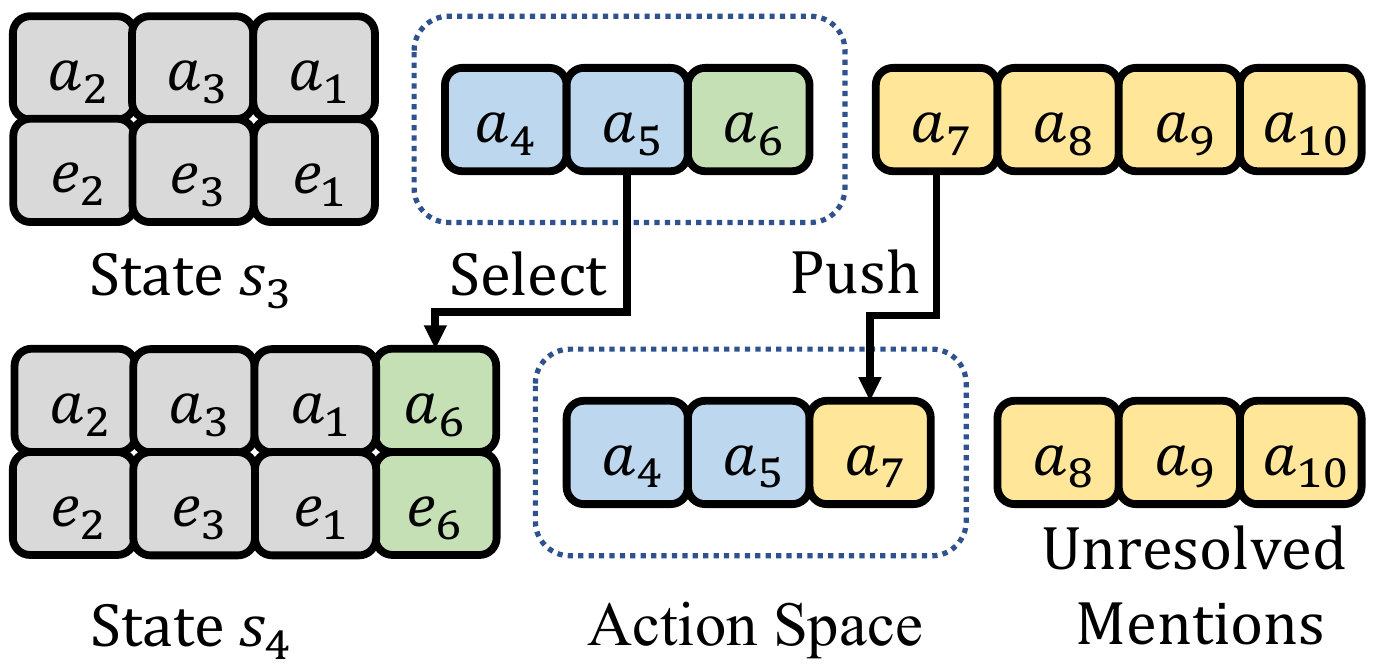}
\caption{An illustration of local fine-tuned action space selection method. The main idea is to use a fixed size sliding window to constantly update the action space.}
\label{finetune} 
\end{figure}

\noindent {\bfseries{Actions.}} The set $A_t$ of possible actions $a_t$ at step $t$ consists of those unresolved mentions. Initially, this set consists of all the mentions, making it difficult for agents to select an appropriate action. Meanwhile, choosing the next action from the above set of actions will greatly disarrange the original order of mentions in the document. We believe the original order of mentions is beneficial to understanding mention and subsequent entity linking. Therefore, we design a local fine-tuned action space selection method, which considers maintaining contextual coherence and reducing action space. The method is shown in Figure \ref{finetune}, the subscript $k$ of $a_k$ refers to the position of mention in the text.

\noindent {\bfseries{Policy Network.}}
After defining the state and action, we can select an action from the action space through a policy network $\pi(a|s)$. We apply a neural attention mechanism to the policy network. Formally, we calculate the relevance score of each element in the state to the action space.
\begin{equation}
c(s^i_t) = \max_{a_t \in A_t}\mathbf{D}_{a_t}^\top \mathbf{B}_3 s^i_t
\end{equation}
\begin{equation}
\mathbf{D}_{a_t} = \sum^n_{j=1}\Psi_{a_t}^j[a_t;\mathbf{e}^j_t]
\end{equation}
where $s^i_t$ is the $i+1$-th element in the state $S_t$. $\mathbf{D}_{a_t}$ is the concatenation of the mention and the candidate entities. $\textbf{e}^j_t$ refers to $j$-th candidate entity that corresponds to mention $a_t$. $\Psi_{a_t}^j$ is the local score of $j$-th candidate entity. $\mathbf{B}_3$ is a parameterized diagonal matrix. We choose Top $K$ elements in $S_t$ as evidence to select the subsequent target mention.
\begin{equation}
{c}(\hat{s}_t) = {\rm Top K}\{c(s^0_t),c(s^1_t),...,c(s^t_t)\}
\end{equation}
Then, the attention weights are as follows
\begin{equation}
w(\hat{s}^j_{t})=\frac{\exp [{c}(\hat{s}^j_{t})]}{\sum_{\hat{s}^k_{t} \in \hat{s}_t} \exp [{c}(\hat{s}^{k}_t)]}
\end{equation}
Finally, we get the probability distribution of each action.
\begin{equation}
\pi_{\theta}(a_t|S_t)={\rm Softmax}( \sum_{\hat{s}^j_{t} \in \hat{s}_{t}} w(\hat{s}^j_{t}) \mathbf{D}^\top_{a_t} \mathbf{B}_4 \hat{s}^j_{t})
\end{equation}
where $\mathbf{B}_4$ is a learnable diagonal matrix. Next, we need to design the reward function to guide the learning of policy for mention prediction.

\noindent {\bfseries{Rewards.}}
In our framework, since one mention selection has a long-term impact on subsequent entity linking, we use a delayed reward to improve the overall performance. In this paper, we will design the reward functions from different perspectives to explore the influence of the order of mention on the subsequent entity selection. When designing the reward function, we follow the principle that if the current motion is linked to the correct entity, we think that the previous information is enough to disambiguate the current motion; if not, we think that the previous information is not enough to disambiguate the current motion, and we need to choose a new mention to disambiguate. In the end, we will get an \textbf{ideal sequence}: all the mentions in the front of the sequence are correctly linked, and there are a few incorrect linked mentions in the back of the sequence, which need other information to disambiguate.
\begin{figure}[ht] 
\centering
\includegraphics[width=3cm]{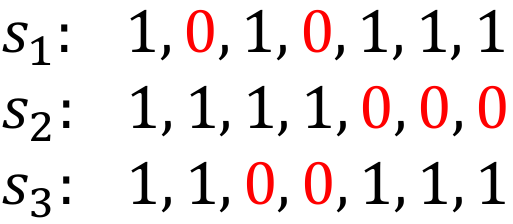}
\caption{An example of the reward functions. $s_1$, $s_2$, and $s_3$ are  different predictions  for  the same sequence, where `1' means the link is correct and `0' means the link is wrong.}
\label{reward_exam} 
\end{figure}

(1) The first reward function ${\rm R1}$ is as follows,
\begin{equation}
{\rm R1}(t) = \frac{\gamma^{L-t}}{L}(-L + I_{\rm first\_error})
\end{equation}
where $\gamma$ is the discount factor. $L$ is the number of mentions, and $I_{\rm first\_error}$  is the index of the first mention with the incorrect linking in the mention sequence.  For example, in Figure \ref{reward_exam}, we can calculate $R1(T)=(-L+I_{first\_error})$, and we get $R1(T)_{s_1}=-6$, $R1(T)_{s_2}=-3$, and $R1(T)_{s_3}=-5$. Finally, the rewards for different moments are $R1(t)_{s_2}>R1(t)_{s_3}>R1(t)_{s_1}$. The reward function can guide the policy network to give priority to learning those easy to be linked mentions.

(2) The second reward function ${\rm R2}$ is designed by adjusting the transition reward. We can get the transition reward $\lambda_{\tau-1,\tau}$ between mentions based on the result of entity linking. For instance, we have a mention sequence $\{m_0,m_1\}$ where the mention has been linked. We select a new mention $m_2$ through the policy network and link it to an entity. If the linking is correct, there will be a large transition reward between $m_1$ and $m_2$. If the linking is wrong, we will get a small transition reward. 
\begin{equation}
{\rm R2}(t) = \frac{\gamma^{L-t}}{L}\sum^L_{\tau=1}\lambda_{\tau-1,\tau}
\end{equation}
where $\lambda_{\tau-1,\tau}$ is the transition reward between action $a_{\tau-1}$ and $a_{\tau}$, where $\lambda_{\tau-1,\tau} \in \{\lambda_{T,T},\lambda_{T,F},\lambda_{F,F},\lambda_{F,T}\}$.  
$\lambda_{T,F}$ refers to the transition reward of situations where the previous entity prediction is correct, and the next prediction is wrong. 'T' and 'F' are True and False, respectively. For example, for $s_2$ in Figure \ref{reward_exam}, we get $R2(t)_{s_2} =\frac{\gamma^{L-t}}{L}(3\lambda_{T,T}+1\lambda_{T,F}+2\lambda_{F,F}) $. This reward function reflects the dynamic selection effect as a whole. If there are more wrong links, the reward function will be smaller, indicating that some mention may not make use of reasonable information. 

(3) We also consider both the number of predictions wrong and their position in the sequence to design the reward function ${\rm R3}$.
\begin{equation}
{\rm R3}(t) = \frac{\gamma^{L-t}}{L}\sum_{d \in error}{-1+(I_d-L)/L}
\end{equation}
where $I_d$ is the index of mention that has the incorrect prediction in the sequence. For the sequences in Figure \ref{reward_exam}, we get $R3(T)_{s_1}=-24/7$,  $R3(T)_{s_2}=-27/7$, and  $R3(T)_{s_3}=-23/7$. Finally, $R1(t)_{s_3}>R1(t)_{s_1}>R1(t)_{s_2}$. This reward function takes link accuracy as the primary optimization objective ($R1(t)_{s_3}>R1(t)_{s_2}$ and $R1(t)_{s_1}>R1(t)_{s_2}$). In the case of the same accuracy, the model is optimized in the direction of constructing the \textbf{ideal sequence} ($R1(t)_{s_3}>R1(t)_{s_1}$).
\subsubsection{Candidate Entity Selection} \label{Candidate Entity Selection} Our dynamic mention selection methods can be transplanted to all entity linking models based on sequential decisions as the basis for subsequent candidate entity selection modules.
Here, we adopt the candidate entity selection method in \citet{yang_learning_2019}, which is also a neural attention method. 
\begin{equation}
\Phi\left(e_{i}, \hat{C}_{i}\right)=\mathbf{e}_{i}^{\top} \mathbf{B}_5 f\left(\hat{C}_t\right)
\end{equation}
where $\hat{C}_t$ denote the list of previously linked entities $\{e_1,e_2,...,e_t\}$,  $f(\cdot)$ is a neural attention function. We first calculate the similarity between each candidate entity and a linked entity $e_i$ in $\hat{C}_t$, and take a maximum value as the score of $e_i$. Then we sort the entities in $\hat{C}_t$ according to their scores and select the Top $K$ entities. Finally, the embedding of these words is weighted and summed according to their scores to get $f(\hat{C}_t)$. We also take mention-entity prior $P(e^j_t|m_t)$ \cite{ganea_deep_2017,le_improving_2018,yang_learning_2019}, the mention-entity type characteristics $\Psi_T(m_t,e^j_t)$ \cite{yang_learning_2019}, and the characteristics $\Phi(e_{t+1},\mathcal{N}(\hat{C}_t))$ \cite{yang_learning_2019} of the neighborhood entities mentioned as the basis for our entity linking, where $P(e^j_t|m_t)$ is the probability of selection $e^j_t$ conditioned on $m_i$, $\mathcal{N}(\hat{C}_t)$ is the neighborhood set of all entities in $\hat{C}_t$ in the knowledge graph. We concatenate the above four features and local feature $\Psi_{m_t}$ as the final feature and input the final feature into a feed-forward network to get the final probability of candidate entities $\hat{P}_\theta(e^i_t|m_t)$.
\subsection{Model Learning}
In the dynamic mention selection module, we get the $R(t)$, which is the expected reward at time step $t$. Our goal is to maximize the expected reward over the mention sequence: 
\begin{equation}
J_{ms}(\theta) = \sum_t\sum_{a_t}\pi_\theta(a_t|S_t)R(t)
\end{equation}
We use the REINFORCE algorithm \cite{sutton2011reinforcement} to optimize the objective function $J(\theta)$.
\begin{equation}
\theta \leftarrow \theta+\alpha\sum_tR(t)\nabla_\theta \log{\pi_\theta(a_t|S_t)}
\end{equation}
In the entity selection stage, we adopt the supervised learning models for the linking model:
\begin{equation}
\begin{aligned}
L_{es}(\theta) &=\sum_{D \in \mathcal{D}} \sum_{m_{t} \in D} \sum_{e \in C_{t}} h\left(m_{t}, e\right) \\
h\left(m_{t}, e\right) &=\max (0, \beta-\hat{P}_\theta(e^*_t|m_t)+\hat{P}_\theta(e^i_t|m_t))
\end{aligned}
\end{equation}
where $\theta$ is the model parameter, $e^*_t$ is the ground truth entity. Finally, we get the final loss by summing the individual loss:
\begin{equation}
L(\theta) = L_{es}(\theta) + \gamma_1 |J_{ms}(\theta)|
\end{equation}
where $\gamma_1 > 0$. In the training stage,  the entities in state $S_t$ and $\hat{C}_t$ are all ground truth entities, which can ensure that the historical information is correct. 
\section{Experiment}
We conduct experiments on the following datasets: (1) For in-domain scenario, we use AIDA-CoNLL dataset \cite{hoffart2011robust}: this dataset contains AIDA-train for training, AIDA-A for validation, and AIDA-B (AI) for testing; (2) For out-domain scenario, we trained on AIDA-train and evaluated on five popular test sets: MSNBC (MS), AQUAINT (AQ), ACE2004 (AC), which were cleaned and updated by \citet{guo2018robust}, CWEB (CW) and WIKI (WI) which were automatically extracted from ClueWeb and Wikipedia respectively. The statistics of these datasets are shown in Table \ref{table1}. `Men' is the number of mentions in all the documents. `Doc' is the number of documents. `Pre-doc' is the average number of mentions in each document. `Gold recall' is the percentage of mentions for which the candidate entities contain the ground truth entity.
\begin{table}[ht]
    \centering
\setlength{\tabcolsep}{2mm}{    
\begin{tabular}{lrrrr}
    \toprule[1.5pt]
     Dataset & Men & Doc & Pre-doc&Gold recall\\
    \hline
     AIDA-train & 18448 & 946 & 19.5 &- \\
    AIDA-A & 4791 & 216 &22.1& 97.2 \\
    AIDA-B & 4485 & 231 &19.4& 98.5 \\
    \hline
    MSNBC & 656 & 20 & 32.8&98.5 \\
    AQUAINT & 727 & 50 & 14.5&93.9 \\
    ACE2004 & 257 & 36 &7.1&91.4 \\
    CWEB & 11154& 320 & 34.8&91.3 \\
    WIKI & 6821 & 320 &21.3&92.6 \\
    \bottomrule[1.5pt]
\end{tabular}}
\caption{Statistics of datasets used in experiments. }
\label{table1}
\end{table}
\subsection{Experimental Settings}
We set the dimensions of word embedding to 300 and used GloVe embedding \cite{pennington2014glove}. The entity embedding is publicly released by \citet{ganea_deep_2017}. We use the following parameter values: $K=7$, the size of the hidden layer is set to $100$, the discount factor $\gamma = 0.9$. The method of generating candidate entities is equal to previous work \cite{le_improving_2018}. The value range of weight $\gamma_1$ of multi-task learning is [0.001, 0.00075, 0.0005, 0.00025, 0.0001]. In R2, we set $\{\lambda_{T,T},\lambda_{T,F},\lambda_{F,F},\lambda_{F,T}\}$ into the following different parameter sets $\{0,-2,-1,0\}$ and $\{0,-L\hat{P}_{\tau-1,\tau},-L\hat{P}_{\mu-1,\mu},0\}$, where $-\hat{P}_{\tau-1,\tau}$ refers to the probability of the incorrect linked entity. 
In the Transformer-based model, we use $4$ layers of the encoder with $6$ attention heads, the vector size is $50$ in self-attention, and the hidden size and feed-forward layer size are $300$ and $[300,600]$. The range of the sliding window is set to $[2, 3, 4, 5, 6, 7, L]$. We use the Adam \cite{kingma2014adam} to train our model for 300 epochs with a learning rate of 2e-4 until validation accuracy exceeds 90\%, afterward setting it to 1e-4.

\subsection{Experimental Results}
\begin{table}[ht]
    \centering
\setlength{\tabcolsep}{3mm}{    
\begin{tabular}{lc}
    \toprule[1.5pt]
     Methods & AIDA-B\\
    \hline
    Deep-ED \citep{ganea_deep_2017} & 92.22  \\
    FGS2EE \cite{hou_improving_2020}& 92.63\\
    DGCN \cite{wu2020dynamic}& 93.13 \\
    BERT-Ent-Sim \cite{chen2020improving} & 93.54 \\
    DRL \cite{fang_joint_2019}& 94.30\\
    DCA-SL \cite{yang_learning_2019}& 94.64 \\
    Attn-global \cite{le_improving_2018} & 93.07 \\
    \hline
    \hline
    Tran-global (300) & 93.42$\pm$0.20\\
    Tran-global (600) & 93.75$\pm$0.17\\
    \hline
    Tran-DyMen (R1)     & 94.26$\pm$0.10    \\
    Tran-DyMen (R2-1)   & 94.15$\pm$0.14    \\
    Tran-DyMen (R2-2)   & 93.18$\pm$0.20    \\ 
    Tran-DyMen (R3)     & 93.30$\pm$0.10    \\
    \hline
    Attn-DyMen (R1)     & \textbf{94.90$\pm$0.10}    \\
    Attn-DyMen (R2-1)   & 94.82$\pm$0.05    \\
    Attn-DyMen (R2-2)   & 94.80$\pm$0.04    \\
    Attn-DyMen (R3)     & \textbf{94.90$\pm$0.10}    \\
    \bottomrule[1.5pt]
\end{tabular}}
\caption{F1 scores (\%) on AIDA-B.}
\label{table2}
\end{table}


\begin{table*}[t]
    \centering
\setlength{\tabcolsep}{2mm}{    
\begin{tabular}{lccccc}
    \toprule[1.5pt]
     System & MSNBC & AQUAINT & ACE2004 & CWEB & WIKI\\
    \hline
    Deep-ED \cite{ganea_deep_2017}                     & 93.70  & 88.50  & 88.50  & 77.90  & 77.50 \\
    FGS2EE \cite{hou_improving_2020}                     & 94.26  & 88.47  & 90.70  & 77.41  & 77.66 \\
    DGCN \cite{wu2020dynamic}                     & 92.50  & 89.40  & 90.60  & \textbf{81.20}  & 77.60 \\
    BERT-Ent-Sim \cite{chen2020improving}               & 93.40  & \textbf{89.80}  & 88.90  & 77.90  & 80.10 \\
    DRL    \cite{fang_joint_2019}                     & 92.80  & 87.50  & 91.20  & 78.50  & \textbf{82.80} \\
    DCA-SL  \cite{yang_learning_2019}                    & 94.57  & 87.38  & 89.44  & 73.47  & 78.16 \\
    Attn-global  \cite{le_improving_2018}               & 93.90  & 88.30  & 89.90  & 77.50  & 78.00 \\
    \hline
    \hline
    Tran-global               & 94.10$\pm$0.16  & 89.45$\pm$0.50  & {90.95$\pm$0.40}  & 77.03$\pm$0.35  & 76.14$\pm$0.30 \\
    \hline
    Tran-DyMen (R1)          & 94.41$\pm$0.10  & 88.13$\pm$0.20  & 90.14$\pm$0.40  & 74.07$\pm$0.10  & 75.95$\pm$0.12 \\
    Tran-DyMen (R2-1)        & 94.87$\pm$0.15  & 88.19$\pm$0.22  & 90.14$\pm$0.40  & 74.15$\pm$0.20  & 75.46$\pm$0.23 \\
    Tran-DyMen (R2-2)        & 94.41$\pm$0.15  & 87.85$\pm$0.20  & 90.94$\pm$0.40  & 73.23$\pm$0.23  & 75.57$\pm$0.21 \\
    Tran-DyMen (R3)          & 94.41$\pm$0.15  & 88.29$\pm$0.10  & 90.54$\pm$0.40  & 73.27$\pm$0.20  & 75.46$\pm$0.32 \\
    \hline
    Attn-DyMen (R1)     & \textbf{95.07$\pm$0.20}  & 87.44$\pm$0.50  & \textbf{91.35$\pm$0.00}  & 75.07$\pm$0.20  & 78.24$\pm$0.20 \\
    Attn-DyMen (R2-1)   & {94.92$\pm$0.29}  & 87.83$\pm$0.23  & 90.54$\pm$0.40  & 74.80$\pm$0.27  & 78.02$\pm$0.32 \\
    Attn-DyMen (R2-2)   & {95.03$\pm$0.15}  & 87.53$\pm$0.34  & \textbf{91.35$\pm$0.40}  & 74.57$\pm$0.17  & 78.06$\pm$0.31 \\
    Attn-DyMen (R3)     & 94.72$\pm$0.15  & 87.72$\pm$0.31  & 90.54$\pm$0.40  & 75.22$\pm$0.30  & 78.27$\pm$0.10 \\
    \bottomrule[1.5pt]
\end{tabular}}
\caption{The performance comparison on cross-domain datasets using F1 score.}
\label{table3}
\end{table*}

Following previous work, we first report the micro F1 scores on AIDA-B and other test sets. Table \ref{table2} shows the results on the AIDA-B. `Attn' is the first local model and `Tran' is the Transformer-based local model. `global' is the global model in \citet{le_improving_2018}. `(300)' means that the feed-forward layer size are $300 $.`R2-1' means we adopt the parameter sets $\{0,-2,-1,0\}$ in the reward function {\rm R2}, and `R2-2' means we adopt the parameter sets $\{0,-L\hat{P}_{\tau-1,\tau},-L\hat{P}_{\mu-1,\mu},0\}$ in the reward function {\rm R2}. We can see that our dynamic mention selection method can achieve the highest performance in-KB accuracy.

Table \ref{table3} shows the results on five cross-domain datasets. It is worth noting that FGS2EE, DGCN, BERT-Ent-Sim, and DRL use the additional text from Wikipedia data for training. In FGS2EE and BERT-Ent-Sim, the global entity linking model is the same as \citet{le_improving_2018}. Comparing Attn-global and Tran-global, our Transformer-based local model has improved performance on MSNBC, AQUAINT, and ACE2004. Our model performs well even when compared to FGS2EE and BERT-Ent-Sim with extra text, especially on ACE2004. Comparing Attn-DyMen and DCA-SL, where the only difference between them is that the order of mention in DCA-SL is fixed, and we can see that the performance of the dynamic mention selection model is improved on all datasets.  At the same time, Tran-DyMen also only performs better on AQUAINT.  

\begin{figure}[t]
\centering
\includegraphics[width=7cm]{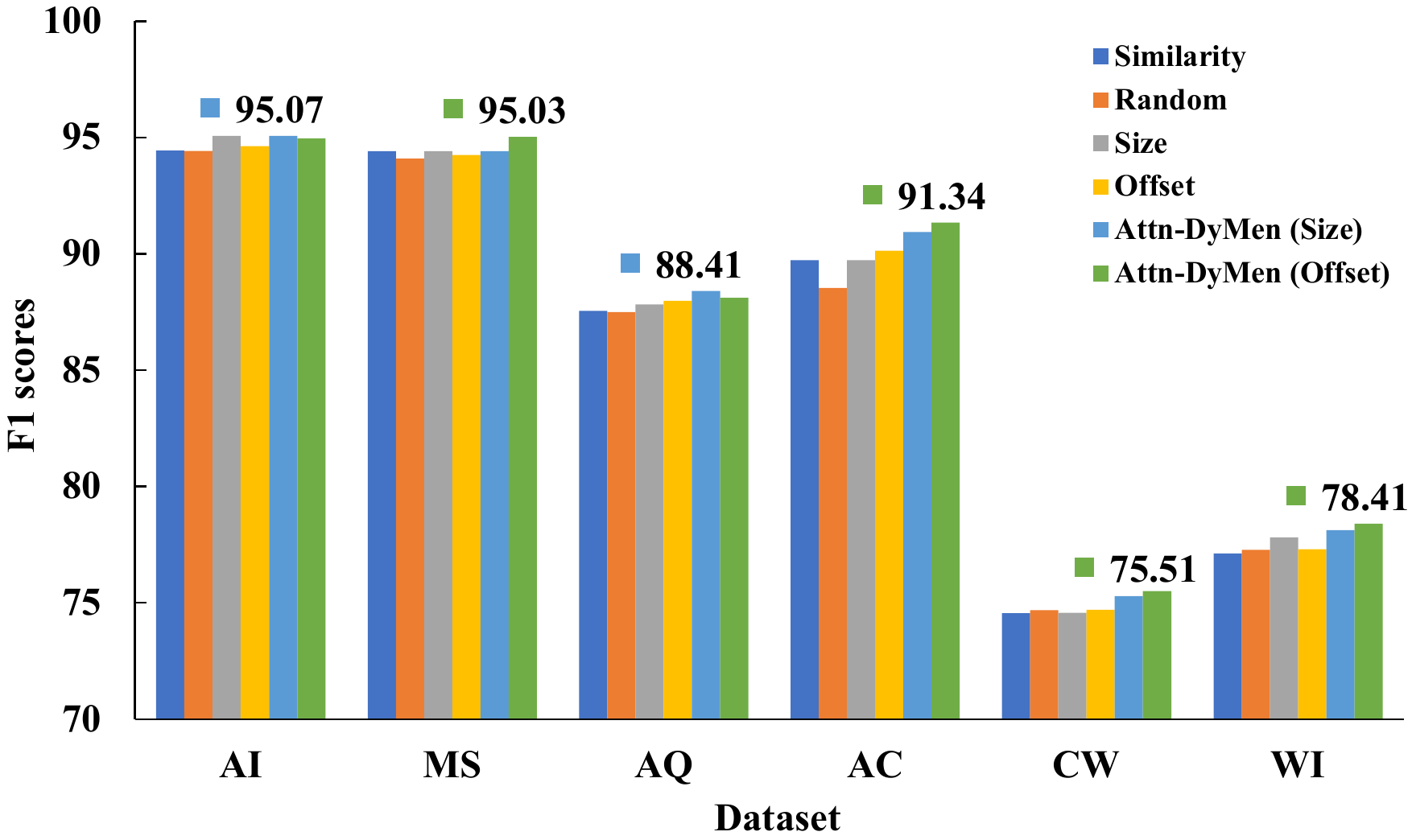}
\caption{The performance comparison of different ranking methods of mention.}
\label{order} 
\end{figure}
\subsection{Experimental Analysis}


\noindent {\bfseries{(1) Order of mention.}}
Figure \ref{order} shows the performance comparison of different ranking methods of mention. `Similarity' refers to sorting by the similarity between mentions. `Random' means to break up the order of the mention in the original document. `Size' first links mentions with smaller candidate sizes, as they tend to be easier to link. `Offset' is the natural order in the original document. `Attn-DyMen (Size)' means we execute our Attn-DyMen model based on the `Size' mention sequence. We can find the results of `Attn-DyMen (Offset)' in Table \ref{table3}. We can see that the dynamic mention selection method is superior to the fixed mention sequence method.

\begin{table}[t]
\setlength{\tabcolsep}{2mm}{ 
\begin{tabular}{c|c|cccccc}
\toprule[1.5pt]
\multicolumn{2}{c|}{$\gamma_1$}  & AI    & MS    & AQ    & AC    & CW    & WI    \\ \hline
{1e-3}   & R1 & 94.7 & 94.6 & 87.6 & 90.1 & 75.5  & 77.5 \\
          & R2 & 94.9 & 94.4 & 87.1 & 90.1 & 75.3 & 78.0    \\ 
          & R3 & \textbf{95.0} & 94.7  & 87.6 & 90.5 & 74.3 & 78.0 \\\hline
{7.5e-4} & R1 & \textbf{95.0} & 94.1  & 87.1 & 90.5 & 75.0 & 77.7 \\
         & R2 & 94.9 & 94.6 & 86.9 & 90.1 & \textbf{75.6} & 77.8 \\ 
         & R3 & {94.8} & 94.9  & 87.0 & 90.5 & 74.7 & 77.7 \\\hline
{5e-4}  & R1 & 94.8 & 94.4 & {87.8} & 90.5 & 74.4 & 77.8 \\
          & R2 & 94.8 & 94.7 & 87.0 & 90.9 & 74.5 & 78.1 \\ \
          & R3 & {94.8} & 94.6  & \textbf{88.1} & 89.7 & 75.0 & 77.9 \\\hline
{2.5e-4} & R1 & 94.9 & \textbf{95.2} & 86.3 &\textbf{91.4} & 74.3 & {78.1} \\
           & R2 & \textbf{95.0} & \textbf{95.2} & 87.1 & 90.9 & 74.1 & 78.0 \\
           & R3 & {94.9} & 94.9  & 87.7 & 90.1 & 74.4 & 77.8 \\\hline
{1e-4}  & R1 & 94.8  & \textbf{95.2} & 86.9 & 90.9 & 74    & 77.8 \\
          & R2 & 94.8 & 95.0 & 87.1 & 90.9 & 74.5  & 78.0 \\
          & R3 & \textbf{95.0} & 94.7  & 87.71 & 90.1 & 74.6 & \textbf{78.3} \\
              \bottomrule[1.5pt]
\end{tabular}}
\caption{The performance comparison of different $\gamma_1$.}
\label{gama}
\end{table}

\noindent {\bfseries{(2) Parameter $\gamma_1$.}}
Table \ref{gama} shows the performance comparison of different $\gamma_1$. We set the size of the window to $4$, and `R2' is R2-2. It can be seen that the parameters have different effects on different reward functions and datasets. For example, in terms of data, MSNBC achieves better performance on smaller $\gamma_1$, while CWEB requires larger parameters $\gamma_1$. In terms of the reward function, we take the average on all datasets, and R1 achieves better performance on larger $\gamma_1$, while R2-2 requires smaller parameters $\gamma_1$.
\begin{figure}[ht] 
\centering
\includegraphics[width=7cm]{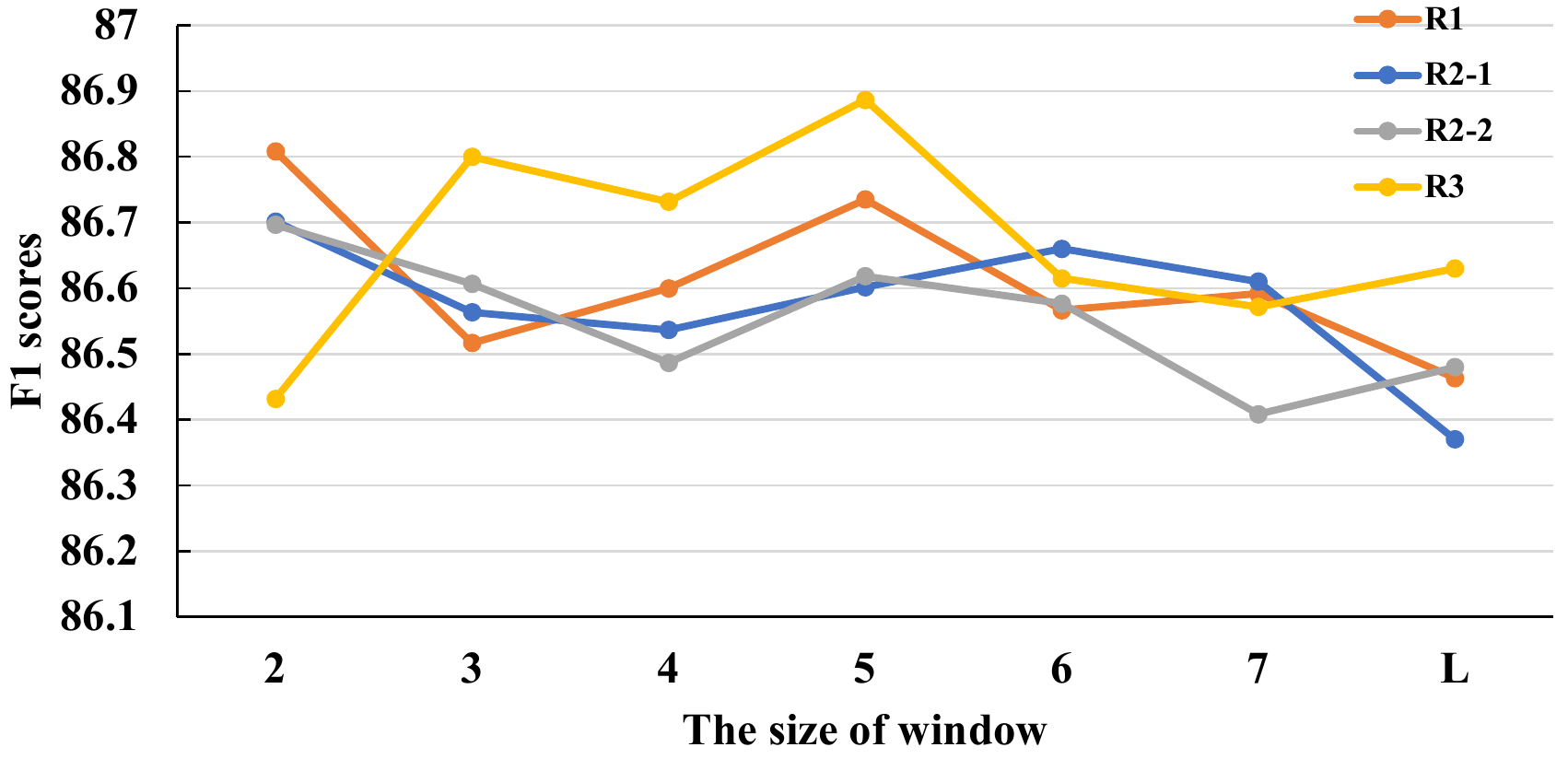}
\caption{Results averaged over all datasets.}
\label{reward} 
\end{figure}

\begin{figure}[ht] 
\centering
\includegraphics[width=7cm]{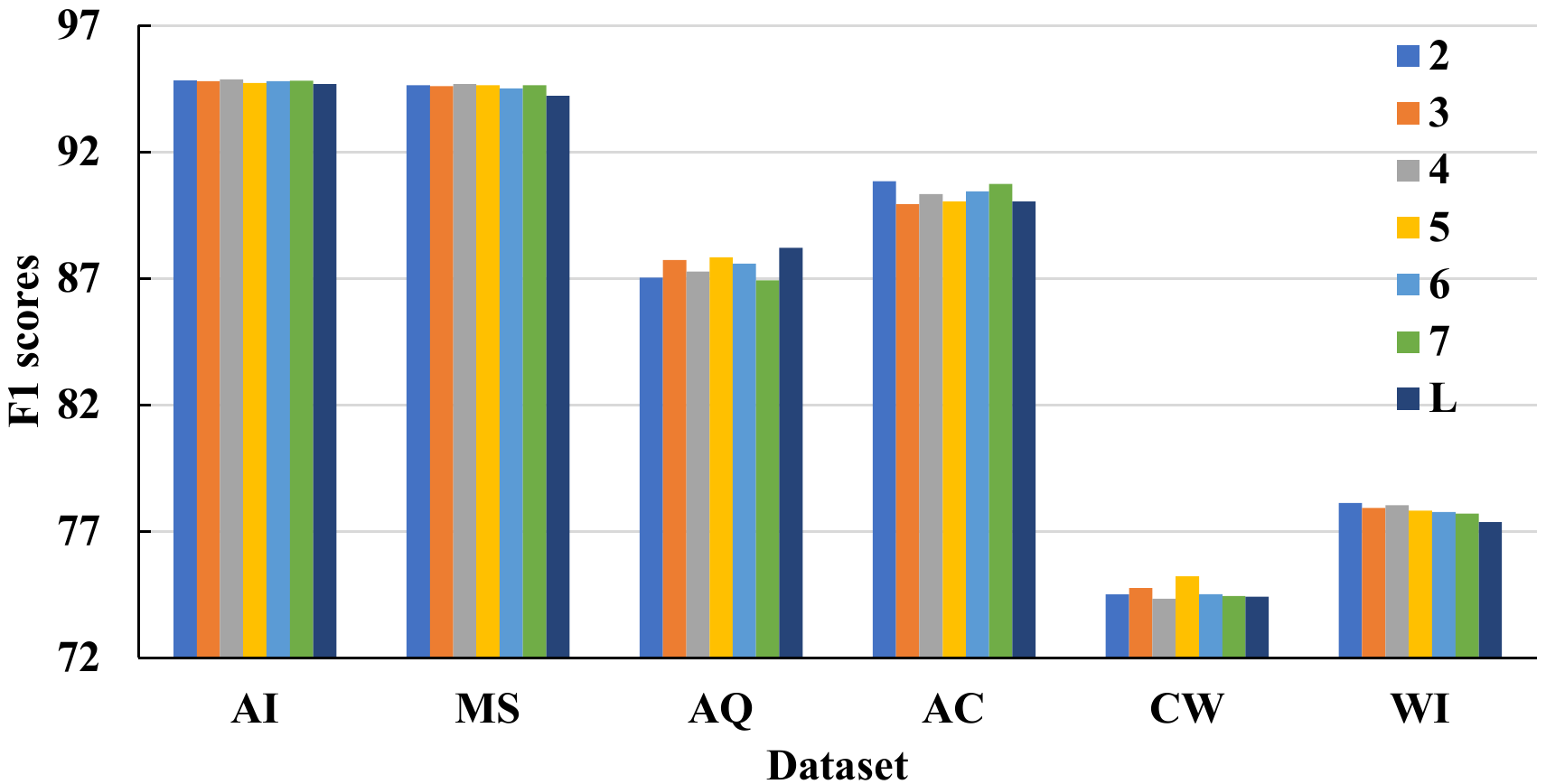}
\caption{Results averaged over all rewards.}
\label{dataset} 
\end{figure}

\noindent {\bfseries{(3) Size of the window.}}
Figure \ref{reward} and \ref{dataset} shows the performance comparison of different window sizes, here $\gamma_1=0.0001$. We can see that different reward functions and datasets have different sensitivity to the size of the window. In general, some datasets require dynamic selection across the entire mention sequence, such as AQUAINT. Other datasets need to select mention within a limited window, such as ACE2004, CWEB, and WIKI. In terms of the reward function, all methods achieve the best performance in the limited window, which shows the effectiveness of the sliding window.

\subsection{Ablation Studies}
\noindent {\bfseries{(1) Transformer-based local model.}} We design ablation experiments on the Transformer-based local model.  It can be observed from Table \ref{tran_abl} that the $N_1$ module are essential for the Transformer-based local model. Compared with Trans-global, position encoding, type encoding, and segment encoding can improve the model significantly.
\begin{table}[ht]
    \centering
\setlength{\tabcolsep}{1.2mm}{    
\begin{tabular}{lcccccc|c}
    \toprule[1.5pt]
     Methods & AI & MS & AQ & AC & CW & WI& Avg.\\
    \hline
    Trans-global & {93.3} & 94.1 &\textbf{90.0}&\textbf{90.9} &76.9 & {75.5} &  \textbf{86.8}\\
    w/o Posi & 93.1& \textbf{94.3} & 89.6&89.7& \textbf{77.7}&75.4  &86.6\\
    w/o Type & {93.6}& 94.1 & 89.2&{90.5}& 77.0&75.2  &86.6\\
    w/o Seg & {93.5}& 94.1 & {89.5}& 90.1& 76.8 & \textbf{75.8}  &86.6\\
    w/o $N_1$ & 74.2& 89.8 & 84.8 & 86.9 & 70.4 & 64.3  &78.4\\
    w/o $N_2$ & \textbf{93.7}& 94.0 & 89.0 &\textbf{90.9}  &76.9 &75.4  &86.7\\

    \bottomrule[1.5pt]
\end{tabular}}
\caption{Ablation studies on Trans-global.}
\label{tran_abl}
\end{table}

\noindent {\bfseries{(2) Feature fusion.}}
To fairly compare the sequential decisions and the pairwise coherence global entity linking method \cite{ganea_deep_2017}, we design ablation studies on feature fusion. We remove the mention-entity type feature $\Phi_T(m_t,e^j_t)$ and the neighborhood entities feature $\Psi(e_{t+1},\mathcal{N}(C_t))$ mentioned in \ref{Candidate Entity Selection}. Table \ref{feature} shows that the pairwise coherence method is superior to the sequential decisions with non-dynamic mention selection \cite{yang_learning_2019} in all datasets. Compared with the pairwise coherence method, the dynamic mention selection method slightly improves some data, but it also has an obvious decline in AIDA-B and WIKI. We think that the reason may be the shortcomings of the sequential decisions itself. When we disambiguate a mention, the sequential method can only be based on the previously linked entities, which may lead to the model can not make full use of all the information in a real sense, especially in the initial stage of the linking, there is little information to utilize. Meanwhile, the existing sequence-based one-way decision methods cannot correct potentially incorrect linked entities. Our approach can alleviate the problem of information inadequacy by adjusting the order of mention and selecting the next target based on historical information. However, it cannot solve the problem caused by the one-way decision.
\begin{table}[ht]
    \centering
\setlength{\tabcolsep}{1.8mm}{    
\begin{tabular}{l|cccccc}
    \toprule[1.5pt]
     Methods & AI & MS & AQ & AC & CW & WI\\
    \hline
    Attn-global          & \textbf{93.1}  & 93.9  & 88.3  & 89.9  & \textbf{77.5}  & \textbf{78.0}\\
    DCA(w/o T-K)         & 91.5  & 93.8  & 87.7  & 89.7  & 76.7  & 77.3 \\
    \hline
    R1 (w/o T-K)        & 92.0 & 94.0  & 88.5 & \textbf{90.1} & 77.1 & 77.2  \\
    R2-1(w/o T-K)      & 92.0  & 94.0 & 89.2 & \textbf{90.1} & 76.6 & 77.0 \\
    R2-2(w/o T-K)      & 91.5  & \textbf{94.1} & \textbf{89.4} & \textbf{90.1} & 77.3 & 76.5 \\
    R3(w/o T-K)        & 92.1  & 93.8  & 88.6 & 89.7 & 77.3 & 77.3 \\
    \bottomrule[1.5pt]
\end{tabular}}
\caption{Ablation studies on feature fusion.}
\label{feature}
\end{table}
\section{Conclusion and Future Work}
This paper introduces a dynamic mention selection global entity linking model. It can dynamically select the next target according to the previously linked entities. Specifically, we design a policy network that can take the linked mentions and entities as historical information and select an appropriate mention from the unresolved mentions to perform entity linking. In selecting mention, we use a sliding window sampling mention to reduce the action sampling space of reinforcement learning and maintain the semantic coherence of mention. Finally, we optimize our model from three different perspectives. Our dynamic mention selection method can be transplanted to all entity linking models based on sequential decisions. Extensive experiments on several public benchmarks demonstrate the effectiveness of our model. In future work, we will design a model to take full advantage of global information between entities and correct for potentially incorrect linked results.

\bibliography{aaai22}

\end{document}